*Review*

# Evolution of Safety Requirements in Industrial Robotics: Comparative Analysis of ISO 10218-1/2 (2011 vs. 2025) and Integration of ISO/TS 15066

**Daniel Hartmann [1*], Kristýna Hamříková [2], Aleš Vysocký [1], Vendula Laciok [2] and Aleš Bernatík [2]**

[1] Faculty of Mechanical Engineering, VSB—Technical University of Ostrava, 708 00 Ostrava, Czech Republic; ales.vysocky@vsb.cz

[2] Faculty of Safety Engineering, VSB—Technical University of Ostrava, 708 00 Ostrava, Czech Republic; kristyna.hamrikova@vsb.cz; vendula.laciok@vsb.cz, ales.bernatik@vsb.cz

**\* Correspondence: daniel.hartmann@vsb.cz**

**Abstract:** Industrial robotics has established itself as an integral component of large-scale manufacturing enterprises. Simultaneously, collaborative robotics is gaining prominence, introducing novel paradigms of human-machine interaction. These advancements have necessitated a comprehensive revision of safety standards, specifically incorporating requirements for cybersecurity and protection against unauthorized access in networked robotic systems. This article presents a comparative analysis of the ISO 10218:2011 and ISO 10218:2025 standards, examining the evolution of their structure, terminology, technical requirements, and annexes. The analysis reveals significant expansions in functional safety and cybersecurity, the introduction of new classifications for robots and collaborative applications, and the normative integration of the technical specification ISO/TS 15066. Consequently, the new edition synthesizes mechanical, functional, and digital safety requirements, establishing a comprehensive framework for the design and operation of modern robotic systems.

## 1. Introduction

ISO 10218-1 [1,2] and ISO 10218-2 [3,4] establish a pivotal global normative framework for the safety of industrial robotics. Classified as Type C standards, they provide detailed specifications for technical and safety requirements tailored to this specific category of machinery. In accordance with the principles of normative hierarchy—as defined, for instance, in ISO 12100 [5]—these standards take precedence over more general Type A and Type B standards in the event of a conflict. Their contribution extends beyond harmonizing requirements for the international market; crucially, they codify a unified approach to risk management across the design, system integration, and operational phases of industrial robotic systems.

The historical evolution of these standards began in 2006 with the publication of ISO 10218-1, which defined requirements for robots as partly completed machinery. However, the comprehensive framework was established in 2011 with the release of the inaugural edition of ISO 10218-2 regarding system integration, alongside a revision of Part 1 [2,4]. This 2011 edition responded to rapid technological advancements in automation by clarifying terminology and functional safety requirements. Nevertheless, it did not address the emerging field of collaborative robotics, which began to penetrate industrial practice in subsequent years. This regulatory gap was bridged by the technical specification ISO/TS 15066:2016 [6], which provided the first comprehensive framework for safe collaborative operation, establishing limits on force, pressure, and speed during human-robot interaction.

The primary objective of the 2025 revision was to address the fundamental transformations that have reshaped the robotics industry over the past decades specifically the proliferation of collaborative applications, the rise of networked systems, and the associated imperatives for cybersecurity and software function validation [9,10]. Furthermore, the new edition strengthens alignment with general standards such as ISO 12100:2010 (General principles for design — Risk assessment and risk reduction) [5] and ISO 20607:2019 (Instruction handbook — General drafting principles) [11], ensuring greater compatibility within the broader machinery safety framework.

This article presents an analytical comparison of the requirements and structural evolution of the ISO 10218-1 and ISO 10218-2 standards (2011 vs. 2025) and delineates the integration of the ISO/TS 15066:2016 into the new edition. The aim is not merely to enumerate individual changes, but to provide a comprehensive overview of the evolving approach to robotic system safety, including a discussion on the impact of these modifications on the design, integration, and certification of robotic workplaces.

The expansions and updates extend beyond cybersecurity to encompass a broad spectrum of thematic areas. Functional safety has been significantly reinforced, with requirements now defined in greater detail and aligned with current versions of ISO 13849-1:2023 (Safety of machinery — Safety-related parts of control systems) and IEC 62061:2021 (Safety of machinery - Functional safety of safety-related control systems). The standard also introduces a novel classification system for robots and applications, distinguishing between different robot classes and collaborative task types to facilitate more precise risk assessment and subsequent certification. The verification and validation domain has been revised to include new methodologies and a clearly structured tabular Annex G, which serves as a systematic tool for assessing compliance with the design and safety requirements for robots and robotic applications detailed in Chapter 5. Furthermore, significant revisions to terminology and marking have harmonized the text with the ISO and IEC frameworks for machinery safety and ergonomics, enhancing cross-document compatibility. Finally, the chapters on information for use have been expanded to incorporate aspects of cybersecurity, software management, emergency procedures, and the validation of safety function parameters.

## 2. Methodology

The methodology of this work is grounded in the comparative content analysis approach [7,8]. This approach was selected for its capacity to systematically evaluate the discrepancies between successive versions of technical and normative documents. The primary objective of the analysis was to systematically classify individual changes, thereby facilitating their structured evaluation and the subsequent interpretation of their impacts. Within the analytical process, all identified changes were systematically categorized into three predefined classes:

1. **Technical changes:** Modifications affecting physical implementation, safety functions, or system parameters. This category encompasses the introduction of new requirements, the tightening or relaxation of limits, as well as the deprecation of specific technical requirements or their replacement with modern procedures. The guiding question for classification in this category is: *Does this change necessitate a different approach to the design, manufacture, or programming of the robot?*

2. **Structural and methodological changes:** Alterations in the organization and philosophy of the document. This includes the reorganization of chapters, the delegation of processes to other standards, changes in the status of annexes, text reduction for clarity, and the standardization of documentation formats. The guiding question for classification in this category is: *Has the location of information within the standard changed, or has the procedural approach been modified?*

3. **Terminological and semantic changes:** Adjustments in definitions and nomenclature that alter or clarify the meaning of the terminology used. This involves the introduction of new terms, shifts in the conceptual understanding of existing terms, and the harmonization of vocabulary with superior standards. The guiding question for classification in this category is: *Has the term used to describe a device or phenomenon changed, or has the meaning of the term itself evolved?*

**Documents used:** The standards selected for comparison were ISO 10218-1:2011 [2], ISO 10218-1:2025 [1], ISO 10218-2:2011 [4], ISO 10218-2:2025 [5], and ISO/TS 15066:2016 [6]. The analysis included only documents that are currently valid or whose validity was terminated by the release of a newer version.

**Analysis procedure**

1. **Study of the new version:** Initially, the new editions of ISO 10218-1 and ISO 10218-2 were studied in detail to comprehend their structure, the logic of their division, and the interrelationships between individual chapters.

2. **Comparison with the previous version:** This phase consisted of a detailed textual comparison. First, a systematic comparison of the 2025 editions with their 2011 predecessors was conducted, analysing each chapter, subchapter, and annex. Concurrently, ISO 10218-2:2025 was subjected to a specific comparison with the technical specification ISO/TS 15066:2016. This additional analytical step was necessary as the new edition of the standard (Part 2) partially incorporates the content of this technical specification. The aim was to precisely identify the mode of integration – i.e., which elements from ISO/TS 15066 were adopted without alteration, which were reformulated, and which were newly defined within the updated standard.

3. **Identification of changes:** This step involved the systematic classification of all identified differences into the three defined categories. For every change in any part of the document, an assessment was made as to whether it constituted a technical, structural, or terminological modification.

4. **Creation of overview tables:** Based on this comparison, tables were prepared summarizing the status of each chapter, indicating whether it remained unchanged, was updated, expanded, or deleted.

5. **Selection of representative changes:** Due to the significant scope of the updates, only a representative selection of changes was included in the article. This selection resulted from a methodological procedure initiated by an expert discussion within the author team, aimed at identifying the most relevant modifications. Only changes meeting the following combination of criteria were included in the final analysis:

   a. **Methodological consistency:** The change had to be clearly classified into one of the three predefined categories.

   b. **Significance of impact:** The relevance of the change to industrial practice and its potential impact on the design, safety, integrity, validation, and certification of robotic systems were assessed.

   c. **Quantitative indication:** As a supplementary criterion, the frequency of occurrence of a given topic across chapters was evaluated.

The interpretation of results was conducted regarding the impacts individual changes may have on key stakeholders in the field of robotics – robot manufacturers, system integrators, and research institutions. Consequently, within the discussion, the identified modifications are analysed not only in terms of their formal normative content but also regarding their practical significance for the design, implementation, and certification of robotic workplaces.

## 3. Comparison of ISO 10218-1:2011 vs. ISO 10218-1:2025

The subject of this chapter is comparative analysis focused exclusively on the first part of the standard ISO 10218-1 [1,2] and its evolution between the 2011 and 2025 editions. The analysis proceeds systematically, comparing individual sections and chapters of both documents. In cases where specific topics are closely interrelated in content across multiple sections, their comparison has been consolidated into a single logical unit to ensure substantive relevance. This approach was selected to maintain consistency with the previously defined analytical methodology.

*3.1. Context and normative framework*

The ISO 10218-1:2025 standard represents the third edition of the foundational document specifying safety requirements for industrial robots, addressed primarily to robot manufacturers. This 2025 edition supersedes the previous ISO 10218-1:2011 (EN ISO 10218-1:2011) and was prepared by Technical Committee ISO/TC 299 in collaboration with CEN/TC 310, in accordance with the Vienna Agreement, which ensures coordination between ISO and CEN. The new standard differs from the previous version primarily in its global character. While the 2011 edition was adopted as a European harmonized standard (EN ISO) and was tightly integrated with the CEN/CENELEC endorsement process and European directives via the informative Annex ZA, the new version functions as an international ISO document without mandatory regional endorsement. This shift from a European to a global ISO framework enhances the exportability of the standard and facilitates its broader application outside the European region.

From a systematic perspective, both editions confirm their classification as Type C standards within the meaning of ISO 12100. As such, they take precedence over general (Type A) or generic (Type B) standards in the event of a conflict. This maintains continuity and binding applicability for all manufacturers of industrial robots who must comply with these requirements.

**Comparison of scope and application**

The scope of both versions of the standard has been significantly expanded and re-fined. The 2011 edition (EN ISO 10218-1:2011) focused primarily on requirements and guidelines for inherent safe design, protective measures, and information for the use of industrial robots. The standard described fundamental hazards associated with robots and established principles for their elimination or risk reduction. Simultaneously, it explicitly stated that it did not address the robot as a complete machine, as the robot is understood as partly completed machinery that becomes a fully functional system only through integration. Among other exclusions, noise emissions were omitted from the scope of the previous EN ISO 10218-1standard, as they were not considered a significant hazard for the robot itself. Furthermore, all non-industrial robots, such as underwater, military, medical, or service robots, were excluded.

In the new edition of ISO 10218-1, this scope is significantly expanded and clarified. The standard explicitly states that it addresses a robot as partly completed machinery intended for use in an industrial environment and provides a detailed list of excluded areas. This list includes, among others, robots:

• used underwater, in law enforcement, military, or space applications;
• medical robots and robots for healthcare, including prosthetics;
• service robots intended for interaction with the public;
• consumer products intended for domestic use;
• equipment for lifting or transporting persons.

The new edition also introduces a list of environments exempted from assessment, such as nuclear facilities, hygienic applications, explosive atmospheres, extreme climatic conditions, underground use, or applications involving mobile platforms. Consequently, the document is clearly delimited to industrial robots used in standard manufacturing environments.

Furthermore, a new formulation appears stating that the standard addresses significant hazards, hazardous situations, and events that are reasonably foreseeable to the integrator during both intended use and reasonably foreseeable misuse. This approach reflects the current concept of risk management according to ISO 12100 and shifts the standard towards a more detailed definition of the context in which robot safety is assessed.

**Normative framework and references**

The introduction to the document also includes a list of cited standards, which has been significantly modernized in the new edition. EN ISO 10218-1:2011 primarily referenced the then-current editions of ISO 13849-1:2006, IEC 62061:2005, ISO 13850, and IEC 60204-1, the new edition lists updated versions of these documents and incorporates additional areas. These include ISO 20607:2019 for the structure of instruction handbooks, ISO 3864 and ISO 7010 for graphical symbols, and ISO 14118:2017 and ISO 14119:2024 for prevention of unexpected start-up and interlocking devices. Conversely, ISO 10218-2 has been removed from the original list of normative references.

**Summary and impact**

A comparison of the two versions indicates that ISO 10218-1 has undergone significant conceptual modernization. The new edition emphasizes international compatibility, an expanded scope, and a clearer definition of industrial applications. The scope has been broadened by a more detailed list of exclusions, the addition of new types of environments, the introduction of the principle of foreseeable misuse, and, crucially, a fundamental update and expansion of the introductory normative references. The most significant changes within this introductory part of the standard are summarized in the following Table 1.

**Table 1** Comparison of context and normative framework of ISO 10218-1 (2011 vs. 2025)

| Area | ISO 10218-1:2011 [2] | ISO 10218-1:2025 [1] | Changes and Implications | Category |
|---|---|---|---|---|
| Scope of the standard | Focused on inherent safe design, protective measures, and information for use. | Confirms the status of partly completed machinery. Newly includes risk assessment arising from reasonably foreseeable misuse. | **Change:** Introduction of the principle of reasonably foreseeable misuse. **Implication:** Tightens requirements for design and risk assessment. | Technical changes |
| Scope exclusions (Application) | Excluded nonindustrial robots (undersea, military, medical, service) and noise emissions. | The list is significantly refined and expanded (e.g., service robots for interaction with the public, consumer products, equipment for lifting people). | **Change:** A much more detailed list of exclusions. **Implication:** Clearer definition of the target group (industrial robots). | Structural and methodological changes |
| Scope Exclusions (Environments) | Not specifically defined. | A list of environments to which the standard does not apply has been newly added (nuclear facilities, hygienic applications, explosive atmospheres, extreme climates). | **Change:** New specification of excluded environments. **Implication:** The standard focuses clearly on standard manufacturing environments. | Structural and methodological changes |

| Normative references | Referenced the then-current versions of standards, e.g., ISO 13849-1:2006, IEC 62061:2005, and ISO 10218-2. | References updated versions of standards. Adds new references: ISO 20607 – instruction handbooks, IEC 62745 – cableless control. Removes ISO 10218-2. | **Change:** Modernization and expansion of the technical framework. **Implication:** Requires the use of modern approaches to the design of control systems, documentation, and interlocking. | Technical changes |

### 3.2. Terms, definitions, and abbreviations

One of the most significant changes in the new edition of ISO 10218-1:2025 is the expansion and systematization of the terminology section. While the previous edition contained only 38 defined terms—comprising 25 main terms and 13 sub terms, the revised version establishes a comprehensive structure of 81 terms, categorized into 14 thematic groups. This restructuring introduces clearer logic, establishes well-defined hierarchical relationships between concepts, and enhances coherence with other horizontal standards, notably ISO 12100:2010, ISO 13849-1:2023, and IEC 62061:2021.

**Scope and thematic focus of terminology**

The original 2011 edition was based on definitions derived from ISO 8373:1994 and focused primarily on fundamental design and safety concepts (e.g., industrial robot, end effector, protective stop, maximum space). In contrast, the new standard extends the terminology to encompass the broader context of robotic systems, applications, and their functional safety. The terminological structure in ISO 10218-1:2025 is now organized into the following areas:

- **Robot, robotic system, application:** Newly defined terms such as collaborative application, collaborative task, robotic cell, and mobile platform reflect the technological shift towards collaborative and mobile robotics.
- **Subassemblies and components:** Technical terms such as axis, auxiliary axis, mechanical interface, manipulator mass, and payload have been clarified.
- **Controls and modes:** Definitions have been expanded to include teach pendant, span of control, simultaneous motion, and the new term, masked message, adopted from IEC 61508-2:2010, which addresses the functional safety of control systems.
- **Programming and software:** Terms such as control program, task program, teaching, and program verification have been harmonized, and notes regarding off-line programming and hand-guided control (HGC) have been added.
- **Functional safety:** This section has undergone fundamental revision and expansion, incorporating new terms such as monitored standstill, software limit, safety function input/output, and validation.
- **Spaces, zones, and distances:** New definitions such as operating space, limiting device, protected space, and separation distance have been introduced, adapting the terminology to requirements for safe zones and collaborative operations.
- **Risk reduction and verification measures:** New concepts such as sensitive protective equipment (SPE) and risk reduction measures in accordance with ISO 12100 have been included, supplemented by verification and validation definitions aligned with ISO 9000:2015.

**Expansion and structuring of concepts**

Compared to the previous version, there has been not only a quantitative increase in the number of definitions but also a qualitative shift in their structural organization. The new standard establishes a terminological system that interconnects concepts based on their logical relationships—ranging from the robotic application through the robotic cell to the user and operator. This approach has improved consistency across the entire ISO 10218 series while aligning the terminology more closely with the needs of industrial practice, where distinguishing between the robot, system, and application levels is critical.

Key innovations also include the incorporation of concepts related to functional and cybersecurity, such as monitored standstill, software limit, and masked message. These terms reflect the growing significance of safety functions implemented within software and network communication interfaces, addressing vulnerabilities highlighted in recent cybersecurity studies [9,10].

**New table of symbols and abbreviations**

The new chapter also introduces a table of symbols and abbreviations, which was entirely absent in the previous edition. This table provides a clear list of abbreviations used throughout the document (e.g., TCP – Tool Centre Point, SPE – Sensitive Protective Equipment, HGC – Hand-Guided Control).

**Summary and impact**

The terminology section of ISO 10218-1:2025 has undergone significant expansion. The new structure and scope of concepts strengthen the link between the design, software, and organizational aspects of safety and aim to contribute to a uniform understanding of concepts across manufacturers, integrators, and users. This shift also impacts the interpretation of subsequent parts of the standard that rely on these definitions, particularly Chapters 5 to 7 regarding functional safety, verification, and information for use. The most significant changes within this part of the standard are summarized in the following Table 2.

**Table 2** Comparison of context and normative framework of ISO 10218-1 (2011 vs. 2025)

| Area | ISO 10218-1:2011 [2] | ISO 10218-1:2025 [1] | Changes and Implications | Category |
|---|---|---|---|---|
| System Hierarchy | Defines industrial robot (3.10), the hierarchy is not linked. | Introduces the following hierarchy: Robot (3.1.1) → Industrial robot (3.1.1.2). | **Change:** Creation of distinct automation levels. **Implication:** Enables precise definition of responsibility in the integration process. | Terminological and semantic changes |
| Symbols and Abbreviations | Chapter 3 contains no summary table of abbreviations. | Newly introduces a table of symbols and abbreviations (e.g., TCP, SPE, HGC). | **Change:** Addition of a missing summary table. **Implication:** Improves technical readability and document clarity. | Structural and methodological changes |
| Collaboration Concept | Defines collaborative operation (3.4) as a state where purposely designed robots work with a human. | Defines the terms collaborative application (3.1.1.6) and collaborative task (3.1.1.7). The term collaborative robot is omitted. | **Change:** Shift from collaborative robot to collaborative application. **Implication:** Emphasizes that safety is determined not by the robot alone, but by the overall risk assessment of the application | Terminological and semantic changes |

| Mobile Robotics | Mentions mobile robots only in Note 3 to definition 3.10 (industrial robot) as manipulating parts of mobile robots. | Introduces the term mobile platform (3.1.2.8) and integrates it directly into the definition of industrial robot (3.1.1.2). | **Change:** Formal terminological anchoring of mobile platforms within the standard. **Implication:** Extends the scope of the standard to mobile platforms combined with manipulators. | Terminological and semantic changes |
|---|---|---|---|---|
| Functional Safety | Defines specific safety functions in Section 3.19, e.g., safety-rated monitored stop (3.19.6) and safety-rated soft axis and space limiting (3.19.3). | Introduces Section 3.1.8, e.g., monitored standstill (3.1.8.4), monitored speed (3.1.8.5), safety function input/output (3.1.8.9/10). | **Change:** Introduction of a set of terms for software-controlled safety functions. **Implication:** Reflects the potential transition from physical stops to software-monitored limits. | Terminological and semantic changes |
| Space Definitions | Defines maximum space (3.24.1), restricted space (3.24.2), and safeguarded space (3.24.3). | Renames safeguarded space to protected space (3.1.9.5). Adds new term separation distance (3.1.9.6). | **Change:** Expansion and refinement of space definitions relevant to protective measures. **Implication:** Original definitions, designed primarily for static physical perimeter guarding, were insufficient for new applications. | Terminological and semantic changes |
| Cybersecurity | Concepts related to cybersecurity is entirely absent. | Introduces the new term masked message (3.1.3.12) with a direct reference to IEC 61508 (communication safety). | **Change:** First introduction of terminology related to data communication security. **Implication:** Reflects the growing interconnection of robot control systems and the need to address risks arising from spoofing or masking. | Terminological and semantic changes |
| Validation and Verification | Defines only user (3.27) as the entity using the robot. | Expands roles in Section 3.1.7: operator (3.1.7.2) – person performing tasks, and user (3.1.7.3) – entity responsible for operation. | **Change:** Division and refinement of roles associated with the robotic system. **Implication:** Enables clearer assignment of responsibilities for training, maintenance (operator) versus overall risk assessment and operation (user). | Terminological and semantic changes |

### 3.3. Risk assessment, design, and risk reduction measures

This subchapter synthesizes the analysis of two distinct chapters of the standard. It is important to note that this structural grouping does not compromise the established methodology or the integrity of the analysis; each chapter of the standard was subjected to a separate comparative assessment to ensure accuracy and detail.

**Risk assessment**

Chapter 4, entitled "Hazard identification and risk assessment" in the 2011 edition, has undergone a reduction in content and a shift in methodological approach in the 2025 revision. The original text contained six paragraphs providing relatively prescriptive guidance. It not only required analysis and assessment to be carried out but also specified an explicit list of six areas that had to be considered – including specific life cycle phases (teaching, maintenance, setting, cleaning) and specific scenarios (unexpected start-up, personnel access). The new version, now simply entitled "Risk assessment," summarizes this chapter in a single concise normative requirement, which states that a risk assessment must be performed for the robot in accordance with ISO 12100:2010.

This transformation represents a structural shift whereby ISO 10218-1 fully delegates the entire risk management methodology. While in the 2011 edition ISO 12100 served primarily as a supporting document mentioned in a note, in the 2025 edition it becomes the sole binding framework for the entire process. The responsibility for the specific method of performing the analysis and risk assessment is thus shifted from the text of ISO 10218-1 to the general process defined in ISO 12100. This is also associated with a change in the status of Annex A (List of significant hazards); while in 2011 it was understood as a direct input for the analysis, in the 2025 revision it is mentioned only in an informative note. Specific requirements for life cycle phases (maintenance, setting), previously listed in Chapter 4, have not been omitted but have been reintegrated directly into specific technical requirements in Chapter 5 (Design and protective measures) and Chapter 7 (Information for Use), where they are addressed in the context of specific risk reduction measures.

**Design requirements and protective measures to reduce risk**

Chapter 5, which constitutes the core part of the standard, has been expanded to approximately three times its original size in the new edition. In the ISO 10218-1:2011 edition, it was entitled "Design requirements and protective measures" and was divided into 15 sections (5.1–5.15). In the new edition, it has been renamed "Design and risk reduction measures" and now contains over 60 specific provisions, consolidated into ten main subsections (5.1–5.10) and supplemented by numerous tables, notes, and normative references. This expansion reflects the shift from primarily mechanical requirements to an integrated approach to safety, encompassing mechanical, electrical, software, and cyber aspects.

**New chapter structure and expansion of requirements**

The original, rather linearly arranged subchapters 5.1–5.4 ("General," "General requirements," "Actuating controls," " Safety-related control system performance (hardware/software)," and "Robot stopping functions") have been transformed and elaborated in greater detail into a set of subsections 5.1.1–5.1.17, which now specifically address individual design areas. New items include:

- **Materials, mechanical strength, and mechanical design (5.1.2):** Specific requirements for overload tests, dynamic tests, and minimum safety factors have been added.
- **Position maintaining and stored energy (5.1.8 and 5.1.12):** These newly specify the obligation to maintain the manipulator in a safe position in the event of a power failure and explicitly address risks arising from gravity and stored energy.
- **Cybersecurity (5.1.16):** A completely new section requiring a cyber threat assessment, security of communication interfaces, user authentication mechanisms, and implementation of encrypted protocols if the cyber threat assessment determines that a threat may lead to a security risk(s).

- **Robot class (5.1.17):** Introduces a new division of robots into Class I and Class II based on parameters of mass, speed, and force. This classification is directly consistent with the test methodology defined in Annex E. Class I is defined for robots with a manipulator mass ≤ 10 kg, maximum force ≤ 50 N, and speed ≤ 250 mm/s; robots exceeding these parameters are automatically classified as Class II.

In parallel with these new topics, established areas have also been revised – for example, requirements for electrical equipment (now harmonized with IEC 60204-1:2021), electromagnetic compatibility (EMC), or robot motion limitation (5.7).

**Expansion in functional safety**

Significant restructuring has occurred, particularly in sections corresponding to 5.4 through 5.10 in the 2011 edition. The new version consolidates these requirements into an integrated subchapter 5.3 "Safety functions" and subsequent sections 5.4 and 5.5. There is an evident shift from general references to full adoption and integration of the current editions of horizontal standards, specifically ISO 13849-1:2023 [12] and IEC 62061:2021 [13].

New criteria are also specified for Performance Level (PL) and Safety Integrity Level (SIL), with direct differentiation based on the robot classification (Class I/II). New subsections have also been added:

- **Parameterization of safety functions (5.3.5):** Specifies principles for configuring (setting) software safety parameters, including verification mechanisms such as the generation of checksums and identifiers.
- **Communication (5.3.6):** Defines transmission network categories (1–3) according to IEC 62280 and requirements for countermeasures against data corruption, delay, or message masking.
- **EMC (5.3.7):** Harmonizes requirements for functional safety and electromagnetic compatibility in accordance with ISO 13849-1:2023 or IEC 62061:2021, depending on the chosen normative approach. However, this requirement is strictly linked to functional safety; other (general) EMC standards may be required in parallel.

**New and updated safety functions**

In direct connection with functional safety, section 5.5 "Other safety functions" has also been expanded. New items have been implemented in this section that were completely missing in the 2011 edition, for example:

- Start/restart interlock and reset (5.5.2);
- Monitored-standstill (5.5.5);
- Stopping time limiting (5.5.6);
- Stopping distance limiting (5.5.7).

Control and limitation of robot motion

The original sections 5.7 to 5.12 (2011 edition), dedicated to operating modes, manual control, simultaneous motion, axis limitation, and singularity protection, have again been significantly revised and consolidated in the new edition. The standard (2025) now systematically defines individual operating modes (newly in section 5.2.7), strictly delineating differences between manual, high-speed manual, and automatic modes. Each of these modes now has more precisely defined activation conditions, associated safety restrictions, and specifications of functions that must be active in each mode.

This is followed by the newly formulated safety function "Speed limit(s) monitoring" (5.5.3). This allows continuous monitoring of the speed of robot components and ensures that the specified limit is not exceeded, which is particularly relevant for manual mode or collaborative applications.

A significant innovation is the introduction of software and dynamic motion limiting (5.7.4–5.7.5), replacing the previous terminology "safety-rated soft axis and space limiting" (from 3.19.3 in the 2011 edition). These limitations enable precise definition of the space in which the robot is permitted to move and allow active response to changes in configuration or operating conditions. Software and dynamic limits thus provide a clear framework for controlling robot motion functions using integrated safety systems.

**Integration of collaborative capabilities**

Section 5.10 now includes a summary of safety capabilities for collaborative operation, which were only briefly mentioned in the previous edition. Individual collaboration methodologies – Hand-Guided Control (HGC), Speed and Separation Monitoring (SSM), and Power and Force Limiting (PFL) – are now described as integrated safety functions. Their detailed implementation is then closely linked to the requirements of ISO 10218-2.

**Summary and impact**

A comparison of the two editions indicates that Chapters 4 and especially 5 of ISO 10218-1:2025 represent the most extensive and substantively significant revision of the entire document. While the 2011 version was primarily mechanically oriented, the new edition creates a comprehensive system of requirements, linking all aspects of industrial robot safety. The standard thus unifies the areas of mechanical design, control, software, and cybersecurity into a single framework, reflecting the current concept of industrial automation as a complex system.

The new edition also refines requirements for functional safety, explicitly specifying required levels of PL d and SIL 2 for Class II robots and PL b or SIL 1 for Class I robots (in accordance with ISO 13849-1:2023 and IEC 62061:2021). Simultaneously, there is a formal standardization of digital functions, including parameterization, secure communication, and monitoring—which are now considered fully validated safety elements of the system.

The extensive restructuring of Chapter 5 is also reflected in its detailed internal organization. The standard now addresses topics such as materials, structural stability, thermal risks, TCP settings, and the integration of auxiliary axes in detail. As previously mentioned, other newly added or significantly revised sections include cybersecurity, robot classification (Class I/II), requirements regarding laser equipment, and rules for the parameterization of safety functions. The most significant changes within this part of the standard are clearly summarized in the following Table 3.

**Table 3**. Key changes in Risk Assessment and Design Requirements of ISO 10218-1 (2011 vs. 2025)

| Area | ISO 10218-1:2011 [2] | ISO 10218-1:2025 [1] | Changes and Implications | Category |
|---|---|---|---|---|
| Risk Assessment Methodology (Chapter 4) | Chapter 4 contained guidance with an explicit list of points to consider and referenced Annex A. | Chapter 4 is condensed into a single normative requirement: perform risk assessment in accordance with ISO 12100. | **Change:** Removal of guidance from Chap. 4. **Implication:** Full responsibility is delegated to the horizontal standard ISO 12100. | Structural and methodological changes |
| Scope and Structure | Chapter 5 divided into 15 sections (5.1–5.15), focused primarily on mechanical and electrical aspects. | Chapter 5 restructured into 10 main, but highly granular subsections (5.1–5.10) with over 60 specific provisions. | **Change:** Complete restructuring and approximately threefold quantitative expansion. **Implication:** The standard is significantly more detailed and covers new domains (software, cyber). | Structural and methodological changes |
| Cybersecurity | The topic is entirely absent from the standard. | Introduces new section (5.1.16) requiring cyber threat assessment and implementation of measures. | **Change:** Introduction of the entirely new domain of cybersecurity. **Implication:** Reflects connected systems (Industry 4.0); manufacturers must newly address cyber risks. | Technical changes |
| Robot Classification and PL/SIL | Sets one general requirement for safety systems (PL=d / SIL 2) for all robots (5.4.2) | Introduces classification (5.1.17) into Class I and Class II and assigns them different PL/SIL requirements (5.3.3) and Annex C. | **Change:** Introduction of robot classification and subsequent differentiation of functional safety requirements. **Implication:** Enables less costly safety solutions for smaller Class I robots. | Technical changes |
| Parameterization and Communication | Lacks specific requirements for software configuration and safety communication. | New sections (5.3.5, 5.3.6) define requirements for parameterization (checksum generation) and safe communication (Categories 1–3). | **Change:** Formal standardization of digital aspects of functional safety. **Implication:** Expands the safety concept to include the integrity of software settings and data transmission. | Technical changes |
| Mechanical Strength | Contains only general requirements for component failure (5.2.3). | New section (5.1.2) defines specific safety factors for static (1.25) and dynamic (1.1) tests. | **Change:** Tightening and specification of mechanical design requirements. **Implication:** Increases demands on mechanical design and proof of robot durability, which was previously addressed only generally. | Technical changes |
| Motion Limiting (Software) | Defines safety-rated soft axis and space limiting (5.12.3) as a specific limiting means. | Renames and integrates the concept as software limit (5.7.4) and dynamic limiting (5.7.5) as part of safety functions. | **Change:** Terminological refinement and deeper integration of software limits. **Implication:** More clearly defines software-controlled zones as full-fledged safety functions. | Terminological and semantic changes |
| New Safety Functions | Defines basic functions emergency stop (5.5.2), protective stop (5.5.3), monitored stop (5.10.2). | Adds, for example, start/restart interlock (5.5.2), stopping time limiting (5.5.6), and stopping distance limiting (5.5.7). | **Change:** Expansion of the portfolio of standardized safety functions. **Implication:** Provides manufacturers and integrators with a wider set of validated functions for addressing specific risks. | Technical changes |

*3.4. Verification, validation and instructions for use*

Chapters 6 and 7 have undergone restructuring between the 2011 and 2025 editions of ISO 10218-1, mirroring the extensive revisions seen in previous chapters. The primary objective was to harmonize the approach to verification, validation, and the provision of information for use with the general principles of ISO 12100:2010 and, notably, ISO 20607:2019 (Instruction handbooks for machinery). While the original version relied primarily on a set of recommendations and illustrative examples, the new standard establishes a rigid framework that precisely defines manufacturer responsibilities, required assessment methodologies, and the format of provided information.

**Verification and validation**

Chapter 6 has been significantly condensed and transformed into a strictly normative format. The original text, which offered a broad overview of potential verification methods (e.g., visual inspection, tests, measurements, documentation review), has been replaced by a compact chapter with an unequivocal mandate: the manufacturer must verify and validate all design elements and implement safety functions in strict accordance with the requirements of Chapters 4 and 5.

The original informative table (formerly Annex F), which served as a list of properties to be verified, has been superseded by a new, substantially expanded table in Annex G. This annex now provides a detailed mapping of individual requirements from Chapter 5 to their corresponding verification and validation methods.

**Information for use**

The originally concise structure of Chapter 7, which was divided into three subchapters (General, Instructions handbook, and Marking), has been expanded into a more comprehensive format comprising five sections. Subchapter 7.5, "Instruction handbook," alone now contains eighteen detailed subsections. The chapter comprehensively covers all aspects of information provision—from warning signals, markings, and safety pictograms to detailed instructions regarding installation, operation, maintenance, cybersecurity, and functional safety.

The new version explicitly mandates that all information for use must comply with the structure defined in ISO 20607 and be localized for the target market. A critical addition involves mandatory instructions regarding cybersecurity (7.5.11), which specify the necessity for organizational measures and provide a description of implemented functions designed to prevent unauthorized access to the robot system.

A completely new section dedicated to functional safety (7.5.12) has been incorporated. This section requires a detailed description of all implemented robot safety functions, including their performance parameters (PL, SIL, PFH, DC, HFT) and their relationship to relevant inputs, outputs, and response times. This signifies a paradigm shift from providing general user information to supplying comprehensive technical documentation suitable for conformity assessment or safety function certification.

Further expansions address installation (7.5.4), commissioning and programming (7.5.6), maintenance (7.5.16), and abnormal and emergency situations (7.5.17). These sections underscore the principle that safety is not merely a static design feature but a dynamic outcome of the robot's entire life cycle—from production and integration to operation and servicing.

Chapter 7 also introduces new requirements for markings and pictograms: they must comply with ISO 7010 and the ISO 3864 series, ensuring the international standardization of safety symbols. Where space on the robot is insufficient to display all data, the standard permits the use of digital identifiers (e.g., QR codes) that link to the complete device data sheet.

**Summary and impact**

Chapters 6 and 7 have evolved from supplementary components of the original standard into fundamental pillars of robot safety management. Chapter 6 enforces transparency in verification processes, while Chapter 7 establishes a detailed framework for communicating safety information to users, integrators, and regulatory bodies. Consequently, the new edition elevates ISO 10218-1:2025 from technical regulation to a com-prehensive system for safety management throughout the entire life cycle of robotic equipment. The most significant changes within this part of the standard are summarized in the following Table 4.

**Table 4** Key changes in verification, validation and information for use in ISO 10218-1 (2011 vs. 2025)

| Area | ISO 10218-1:2011 [2] | ISO 10218-1:2025 [1] | Changes and Implications | Category |
|---|---|---|---|---|
| Verification Methodology | Chapter 6.2 provided an informative list of possible verification methods (e.g., visual inspection, tests, measurements). Referenced Annex F as a list of properties to be verified. | Chapter 6 is concise and normative. Annex F is replaced by a more detailed Annex G, which serves as a linking tool for each requirement from Chapter 5. | **Change:** Transition from an informative list of methods to a normative mapping matrix. **Implication:** Increases transparency and enforceability of the verification process. | Structural and methodological changes |
| Structure of Information for Use | Basic division into 3 sections (General 7.1, Instruction handbook 7.2, and Marking 7.3). | Detailed restructuring into 5 main sections (7.1–7.5). Section 7.5 "Instruction handbook" itself has 18 specific subsections (7.5.1–7.5.18). | **Change:** Radical restructuring and granulation of content according to the life cycle. **Implication:** Significantly increases clarity and specificity of documentation requirements. | Structural and methodological changes |
| Framework for Instruction Handbook | References ISO 12100 and IEC 60204-1 General (7.1). Does not contain a reference to a specific standard on drafting instructions. | Explicitly requires compliance with the structure of ISO 20607:2019 and defines requirements for language localization (7.1). | **Change:** Introduction of a normative reference to ISO 20607. **Implication:** Unifies the structure of instruction handbooks across engineering and places a formal requirement on their content and form. | Structural and methodological changes |
| Functional Safety Documentation | Requires general information on "safety-related control system performance of the robot" (7.2). | Introduces a new, highly detailed section (7.5.12) requiring full technical specification of all safety functions (including PL, SIL, PFH, DC, HFT). | **Change:** Shift from general description to a requirement for complete technical specification of parameters. **Implication:** The integrator newly receives data necessary for certification and validation of the entire system. | Technical changes |

| Cybersecurity Documentation | Topic is entirely missing from the standard. | Introduces a new section (7.5.11) requiring a description of implemented functions and the necessity of organizational measures by the user. | **Change:** Introduction of requirements for documenting cybersecurity. **Implication:** Reflects risks of connected systems; transfers the obligation to the manufacturer to inform the user about security measures and residual risks. | Technical changes |
|---|---|---|---|---|
| Safety Pictograms and Signs | Specific standards for pictograms are not specified. | Explicitly requires compliance with ISO 7010 (symbols) and the ISO 3864 series (colours, shapes) to ensure international unification (7.4). | **Change:** Specification and harmonization of requirements for safety symbols. **Implication:** Ensures uniform and comprehensible visual marking across different manufacturers and regions. | Technical changes |
| Form of Marking on the Robot | Requires physical marking of all data on the robot (7.3). | Retains the requirement for physical marking but newly permits the use of digital identifiers (QR code, barcode) if there is insufficient space on the robot (7.3) | **Change:** Permission of machine-readable labels as an alternative for providing complete data. **Implication:** Modernization of the approach to marking, enabling provision of more extensive data. | Technical changes |

## 3.5. ISO 10218-1 Annexes

The annexes of the standard have undergone significant restructuring in the 2025 edition, mirroring the overall shift towards a more detailed and technical conception of the entire document. This transformation involved renaming, deprecation, and, most notably, the addition of a series of new, normative and informative annexes that fundamentally expand and specify the requirements of the main text.

**Changes in existing annexes**

- Annex A (2011) vs. Annex A (2025) (List of significant hazards): While remaining informative, its scope is now explicitly limited to the robot prior to integration. The content has been condensed from 10 to 7 risk groups; hazards related to the environment, equipment, and noise have been removed, as they now fall fully under the purview of integration (ISO 10218-2).
- **Annex B (2011) vs. Annex H (2025) (Stopping time and distance metric):** This normative annex (now designated as Annex H) has undergone a significant change in scope. Whereas in 2011 it was generally applicable, in the 2025 edition its measurement requirements apply exclusively to robots lacking modern safety functions that limit stopping time or distance (in accordance with 5.5.6 and 5.5.7). Furthermore, the measurement scope has been extended to include Category 2 stops, and data is re-quired at 33%, 66%, and 100% of speed/load/extension.
- **Annex F (2011) vs. Annex G (2025) (Means of verification):** Although the validation matrix format has been retained, its content has been substantially expanded from 8 to nearly 19 pages to map the new requirements of Chapter 5 in detail. Minor adjustments were also made to the verification methods themselves (A-F); the original method "F – review of risk assessment in relation to the task at hand" has been modified in the new edition to the more general "F – Review of risk assessment and/or FMEA".

- **Annex E (2011) vs. Annex F (2025) (Labelling):** The informative annex (now Annex F) has been expanded from the original two symbols (modes) to seven, covering additional states such as power off or emergency stop.
- **Annex ZA (Relationship with EU Directive):** This informative annex has been relocated to the beginning of the document, and its format has changed from a general declaration to a specific mapping table (Table ZA.1) linking individual clauses of the standard to the essential requirements of the Machinery Directive (2006/42/EC).

**Removed and new annexes**

The 2025 edition removed certain annexes from 2011, typically because their content was elevated to a normative requirement and integrated into Chapter 5. This applies to Annex C (2011) (3-position enabling device) and Annex D (2011) (Optional features). Features previously considered optional (e.g., programmable safety limits) are now standard components of design requirements. Concurrently, a series of entirely new annexes has been introduced:

- **Annex B (New – Illustration of spaces):** An informative graphical aid visualizing the definitions of spaces (maximum, restricted, protected) from Chapter 3.
- Annex C (New – Functional safety performance): A normative annex defining, in tabular form, all safety functions (mandatory and optional), their triggering events, and intended results.
- **Annex D (New – Functional safety information):** An informative annex providing a template (sample table) for reporting technical data of safety functions (PL, SIL, response times) as required by Chapter 7.
- **Annex E (New – Test methodology for Class I):** A technical normative annex defining the precise test methodology for Maximum Power and Force Measurement (FMPM) for the purpose of classifying a robot as Class I.
- **Annex I (New – Start/restart/reset implementation):** An informative diagram clarifying the implementation of start/restart interlock functions (in accordance with 5.5.2).

**Summary and impact**

The restructuring of the annexes confirms the overarching trend of ISO 10218-1:2025. The annexes no longer serve merely as supplementary material but represent integral tools for the implementation (Annex C), testing (Annex E), and validation (Annex G) of safety requirements. The massive increase in scope (particularly Annex G) and the intro-duction of new normative annexes (C, E) impose significantly higher demands on robot manufacturers regarding conformity assessment and technical documentation. The most significant changes within this part of the standard are summarized in the following Table 5.

**Table 5** Key changes in the Annexes of ISO 10218-1 (2011 vs. 2025)

| Area | ISO 10218-1:2011 [2] | ISO 10218-1:2025 [1] | Changes and Implications | Category |
|---|---|---|---|---|
| List of Significant Hazards (Annex A – informative) | Informative list of 10 risk groups derived from ISO 12100. | Informative list of 7 risk groups, restricted to the robot prior to integration. | **Change:** Content reduction. **Implication:** Delineation of responsibility: manufacturer – robot, integrator – robotic application. | Structural and methodological changes |
| Stopping Time/Distance (Annex B – normative vs. Annex H – normative) | Measurement requirements for all robots. | Applies only to robots without stopping limiting functions. Requires data also for Category 2 stops at 33/66/100% of maximum speed, payload, and extension. | **Change:** Increased measurement detail. **Implication:** The annex is now a methodology for legacy robot types; modern robots with functions per 5.5.6/5.5.7 do not use it. | Technical changes |
| 3-Position Device (Annex C – informative) | Informative annex containing a functional diagram of the 3-position enabling device | The annex has been removed. | **Change:** Removal of informative annex. **Implication:** Requirements for 3-position devices are now fully integrated into Chap. 5 (5.5.4). | Structural and methodological changes |
| Optional Features (Annex D – informative) | Informative list of optional safety features. | The annex has been removed. | **Change:** Removal of the annex on optional features. **Implication:** Most of these features (e.g., software limits) have been incorporated into standard normative requirements. | Structural and methodological changes |
| Symbols (Annex E – informative vs. Annex F – informative) | Contains 2 symbols for operating modes. | Expanded to 7 symbols (incl. power off, emergency stop, external control). | **Change:** Content expansion and redesignation. **Implication:** Better visual standardization of robot states. | Terminological and semantic changes |
| Verification (Annex F – normative vs. Annex G – informative) | Original validation table mapping each requirement from original Chapter 5 (8 pages). | More detailed validation table for each requirement from Chap. 5 (Nearly 19 pages). | **Change:** Change of status from normative to informative. **Implication:** Binding nature is in the main text; the annex provides flexibility in method selection. | Structural and methodological changes |
| Harmonization (Annex ZA – informative) | Informative. General declaration of presumption of conformity with 2006/42/EC. | Relocated to the beginning of the standard. Contains a specific mapping table (ZA.1) between clauses and Directive requirements. | **Change:** Relocation and transformation into a detailed mapping table. **Implication:** Clarifies and facilitates the demonstration of compliance with EU legislation. | Structural and methodological changes |
| New Annexes (B, C, E, I) | New Annexes (B, C, E, I) | Introduced new annexes: B (Informative illustration of spaces), C (Normative list of safety functions), E (Normative Class I test methodology), I (Informative restart logic diagram). | **Change:** Introduction of a set of new technical and informative annexes. **Implication:** The standard provides new tools for understanding (B, I), implementation (C), and testing (E). | Technical changes |

## 4. Comparison of ISO 10218-2:2011 vs. ISO 10218-2:2025

The subject of this chapter is a comparative analysis focused on the second part of the standard ISO 10218-2 [3,4], which specifies the requirements for the integration of robotic applications and cells, and its evolution between the 2011 and 2025 editions. The analysis proceeds systematically, comparing individual sections and chapters of both documents. In cases where specific topics are closely interrelated across multiple sections, their comparison has been consolidated into a single logical unit to ensure substantive relevance. This approach was selected to maintain consistency with the analytical methodology employed in the previous chapter when comparing Part 1.

## 4.1 Context and normative framework

ISO 10218-2:2025 represents the second edition of the pivotal document defining safety requirements targeted at system integrators. It does not address the robot itself (covered in Part 1 for manufacturers) but focuses on its integration into robotic applications and robotic cells. This edition cancels and replaces the previous standard ISO 10218-2:2011 (EN ISO 10218-2:2011) and was, like Part 1, prepared by Technical Committee ISO/TC 299 (Robotics) in collaboration with CEN/TC 310 (Advanced Automation Technologies) in accordance with the Vienna Agreement.

The most significant change compared to the previous version is the integration of most requirements from the technical specification ISO/TS 15066:2016. Consequently, requirements for collaborative applications, previously addressed in a separate (and merely informative) document, become a normative component of the base standard. This is accompanied by a significant terminological shift, the new standard explicitly abandons the previously used terms "collaborative robot" and "collaborative workplace," emphasizing instead that collaboration is exclusively a property of the application, rather than the machine itself.

From a systematic perspective, the standard confirms its classification as a Type C standard within the meaning of ISO 12100. This defines its precedence over general Type A and Type B standards in the event of conflict and establishes direct binding applicability for entities performing integration (integrators) and subsequently for users of robotic applications.

**Comparison of scope and application**

The most significant modification in the new edition is the shift from "robotic system" (2011) to "robotic application" (2025). A robotic application is a broader concept that, unlike a mere system, also encompasses workpieces, the task program, and machines and equipment supporting the application and its intended tasks. This shift more accurately reflects the reality of integration, where risk arises not solely from the system but from the entire application.

The second pivotal point declared in the introduction to the 2025 version is the integration of most requirements from the technical specification ISO/TS 15066:2016. Associated with this is a crucial terminological step: the new standard deliberately deprecates the terms "collaborative robot" and "collaborative operation." It emphasizes that the collaborative nature is not an attribute of the robot, but exclusively a property of the application, which must be designed, verified, and validated as a holistic entity.

The 2025 edition also significantly refines its scope through a substantially expanded list of exclusions in Chapter 1 (Scope). While the 2011 version was relatively general, the 2025 edition explicitly excludes applications in military, medical, aerospace, or domestic environments. From a technical perspective, it is significant that the scope now also excludes mobility in terms of integrating robots into unmanned trucks or mobile platforms.

**Normative framework and references**

Chapter 2 (Normative references) reveals the depth of the technical revision. The normative basis of the 2025 edition is significantly expanded and updated. The update regarding functional safety reflects an expected generational shift. Requirements no longer refer to ISO 13849-1:2006 and IEC 62061:2005, but to their current editions, ISO 13849-1:2023 and IEC 62061:2021. This represents a major technical change in the approach to control system design.

Furthermore, there is a new normative requirement for documentation harmonization through compliance with ISO 20607:2019 (Instruction handbooks), reflecting stricter demands on the quality and structure of accompanying documentation. The list of new normative references itself clearly signals the integration of numerous other topics newly incorporated into the standard. These include, for example, visual safety (the entire ISO 3864 series and ISO 7010 for safety signs and pictograms), ergonomics and access (ISO 15534 series – access openings), specific hazards (ISO 13732 – hot/cold surfaces; IEC 60825-1 – laser safety), and specific components (IEC 60947-5-8 – 3-position enabling de-vices).

**Summary and impact**

An analysis of the introductory chapters and normative references indicates that ISO 10218-2:2025 is not merely a revision but rather a redefinition of the integration standard. The transition from "system" to "application" and the substantial absorption of requirements for collaborative applications (formerly ISO/TS 15066) present the integrator with new, considerably more complex tasks. The expanded normative framework, particularly the update of functional safety standards and the introduction of requirements for instruction handbooks (ISO 20607), signals a significant tightening of requirements for both technical design and accompanying documentation. The most significant changes within this introductory part of the standard are summarized in the following Table 6.

**Table 6** Key changes in the context and normative framework of ISO 10218-2 (2011 vs. 2025)

| Area | ISO 10218-2:2011 [4] | ISO 10218-2:2025 [3] | Changes and Implications | Category |
|---|---|---|---|---|
| Collaboration Concept | References collaborative operation. The topic is not deeply specified in the standard; details were later addressed in a separate Technical Specification. | Integrates most requirements from Technical Specification 15066:2016. Deprecates terms collaborative robot and collaborative operation. | **Change:** Requirements for collaborative applications elevated to part of the standard. **Implication:** Significantly tightens and unifies requirements for design and validation of collaborative applications. | Technical changes |
| Scope of the Standard | Focuses primarily on industrial robot systems (robot + end effector + peripherals). | Focuses primarily on industrial robotic applications (system + workpiece + task program + other machines). | **Change:** Shift from system to application. **Implication:** Expands the scope of the standard, better reflects the reality of integration where risks arise from the entire process. | Structural and methodological changes |

| Definition of Scope (Exclusions) | Lists general exclusions (e.g., noise). | Contains a significantly expanded and specific list of exclusions (e.g., military, medical) | **Change:** Detailed specification and expansion of exclusions. **Implication:** The standard defines its boundaries more clearly. | Structural and methodological changes |
| Documentation Requirements | Generally, states that information for use must be provided. | Introduces a new requirement for compliance with ISO 20607:2019 (Instruction handbook). | **Change:** Introduction of a specific standard for the structure and content of the instruction handbook. **Implication:** Increases formal demands on the quality, structure, and content of accompanying documentation. | Structural and methodological changes |
| Harmonization of Pictograms | Does not contain specific normative references to standards for safety marking. | Newly requires compliance with ISO 7010 (registered signs) and the ISO 3864 series (colours, shapes). | **Change:** Introduction of specific standards. **Implication:** Unifies requirements for used pictograms and warning symbols. | Technical changes |

## 4.2. Terms, definitions, and abbreviations

Chapter 3, dedicated to terminology, has undergone an extensive structural transformation in the new edition of ISO 10218-2. This change is signalled by the very title of the chapter itself: while the 2011 edition bore the concise title "Terms and definitions," the 2025 revision expands it to "Terms, definitions, symbols, and abbreviated terms". Whereas the 2011 version was merely a brief supplement referencing Part 1, the 2025 version becomes a comprehensive and largely self-contained lexicon that absorbs not only the terminology from Part 1 but also technical specifications for collaborative robotics.

**Scope and thematic focus of terminology**

ISO 10218-2:2011 contained only 15 definitions of its own (3.1 to 3.15). For all fundamental concepts (such as robot, system, etc.), it referred strictly to ISO 10218-1 and ISO 12100. The standard was narrowly focused on integration (integration, integrator) and spatial layout (collaborative workspace, distance guard).

ISO 10218-2:2025 is orders of magnitude more extensive in this chapter, as the terminology has been significantly expanded. The chapter first absorbs fundamental concepts from ISO 10218-1:2025, such as industrial robot (3.1.1.2), mobile platform (3.1.2.10), or manipulator (3.1.2.7). It then extends this foundation with specific integration concepts (adopted from the 2011 version) and adds several new definitions, for example, software limit (3.1.8.8). Consequently, the standard becomes a semantically self-sufficient and independently usable document for integrators.

**Expansion and structuring of concepts**

A qualitative shift occurs in the concepts that the standard adds beyond the scope of Part 1. The 2011 edition defined a collaborative robot (3.2) as a specific type of machine. The 2025 edition abandons this concept and introduces detailed terminology for collaborative applications:

- **Application hierarchy:** A clear sequence is defined: robotic application (3.1.1.4) → collaborative application (3.1.1.6) → collaborative task (3.1.1.7).

- **Roles:** The roles of integrator (3.1.7.2), operator (3.1.7.3), and user (3.1.7.4) are clarified.
- **Biomechanics:** The entire section 3.1.12 is dedicated to human-robot contact. It introduces fundamental concepts for Power and Force Limiting (PFL) methods, such as quasi-static contact (3.1.12.3) and transient contact (3.1.12.4).
- **Sensors and spaces:** Concepts necessary for the integration of protective devices are added, e.g., detection zone (3.1.9.9) and separation distance (3.1.9.10).

New table of symbols and abbreviations

The 2025 edition introduces a new section 3.2, "Abbreviated terms and symbols." This table was completely absent in the 2011 edition. It now provides an extensive and clear list of dozens of abbreviations used in the standard (e.g., PFL, SSM, HGC, SPE, PL, SIL, TCP), significantly improving the technical readability and unambiguity of the document.

**Summary and impact**

The terminology chapter in ISO 10218-2:2025 has undergone a significant transformation. By duplicating terminology from Part 1 while simultaneously adding highly specific definitions from the field of integration and collaborative safety, Part 2 becomes the primary and largely self-sufficient reference document for integrators. Integrators no longer need to consult other standards for definitions of basic concepts or to understand the biomechanics of collaborative contact. The most significant changes within this part of the standard are summarized in the following Table 7.

**Table 7** Evolution of key terms, definitions, and structural hierarchy in ISO 10218-2 (2011 vs. 2025)

| Area | ISO 10218-2:2011 [4] | ISO 10218-2:2025 [3] | Changes and Implications | Category |
|---|---|---|---|---|
| Scope and Self-sufficiency | Very concise chapter (15 definitions). Refers primarily to Part 1 and ISO 12100 For all basic concepts. | Chapter content significantly expanded (>80 definitions). Duplicates terminology from Part 1 (2025) and introduces new terms. | **Change:** Shift from a brief supplement to a fully self-sufficient terminological standard. **Implication:** The integrator is no longer required to cross-reference Part 1. | Structural and methodological changes |
| Hierarchy and Structure of Terms | Terms listed in a linear, flat list; terms such as application, integrator, cell defined independently without indication of mutual interconnection. | Introduces a thematic and hierarchical structure. Main concepts (e.g., 3.1.1 "Robot...", 3.1.1 "Industrial environment") serve as superordinate categories under which related terms are logically grouped. | **Change:** Transition from a linear list to a semantically structured (nested) ontology. **Implication:** Improved orientation in terminology; the hierarchy of concepts is immediately apparent. | Terminological and semantic changes |
| Collaboration Concept | Defines collaborative robot (3.2) as a specific type of machine designed for collaboration. | Deprecates the term collaborative robot. Introduces hierarchy collaborative application (3.1.1.6) and collaborative task (3.1.1.7). | **Change:** Shift from machine (robot) to process (application/task). **Implication:** Emphasizes that safety is determined not by the robot alone, but by the overall risk assessment of the application. | Structural and methodological changes |
| Contact Biomechanics | Terminology for describing human-robot collision is entirely absent. | Introducing new section 3.1.12 defining | **Change:** Introduction of biomechanical terms to describe collision characteristics. | Terminological and semantic changes |

| | | quasi-static contact (3.1.12.3) and transient contact (3.1.12.4). | **Implication:** Provides the necessary basis for PFL collaboration methods requiring quantification and measurement of contact forces. | |
| --- | --- | --- | --- | --- |
| Symbols and Abbreviations | Chapter 3 contains no summary table of abbreviations. | Newly introduces a table of symbols and abbreviations (e.g., TCP, SPE, HGC). | **Change:** Addition of a missing summary table. **Implication:** Improves technical readability and document clarity. | Structural and methodological changes |
| Role Definitions | Defines only integrator (3.7) and application (3.1). | Detailed specification and differentiation of roles: integrator (3.1.7.2), operator (3.1.7.3), and user (3.1.7.4). | **Change:** Refinement and expansion of definitions of human roles in the life cycle. **Implication:** Enables more precise addressing of requirements and responsibilities. | Terminological and semantic changes |
| Mobile Robotics | Terminology focuses exclusively on stationary robots. Mobile platforms are not listed in the standard. | Introduces the term mobile platform (3.1.2.10) and integrates it directly into the definition of industrial robot (3.1.1.2). | **Change:** Terminological anchoring of mobile platforms within the standard. **Implication:** Extends the scope of the standard to mobile platforms combined with manipulators. | Terminological and semantic changes |

*4.3. Risk assessment*

Chapter 4, which defines the methodological foundation for the entire standard, has undergone significant reform. This transformation is evident in the title change itself, shifting from the prescriptive "Hazard identification and risk assessment" (2011) to the overarching and process oriented "Risk assessment" (2025). The 2011 edition functioned as a standalone guide containing principles, processes, and specific checklists. Conversely, the 2025 edition fully delegates the general risk assessment process to ISO 12100, focusing instead on the detailed specification of inputs and characteristics unique to robotic applications, including collaborative systems.

**Methodological shift and process delegation**

The most significant structural change is the abandonment of the model wherein ISO 10218-2 duplicated or paraphrased the principles of ISO 12100.

- **2011 Version:** This version was highly prescriptive, containing separate sections describing risk reduction principles (4.1), defining the risk assessment process (4.3), and detailing its individual steps, such as determining limits (4.3.2) or identifying hazards (4.4).

- **2025 Version:** This version adopts a strategy of delegation. The introductory section 4.1 establishes a single normative requirement: " risk assessment must be performed for the robot in accordance with ISO 12100:2010. Consequently, all methodological steps (identification, estimation, evaluation) are fully transferred to ISO 12100:2010. The remainder of Chapter 4.1 consists solely of informative notes referencing tools (e.g., ISO/TR 14121-2) and other annexes of the standard (A, C, E).

**Input specification: Layout design and task identification**

While the process itself has been delegated, the requirements for inputs into this pro-cess have been expanded and refined. The 2011 version included an extensive section 4.2, "Layout design," which functioned as a checklist for the integrator (covering physical boundaries, access, ergonomics, etc.), and section 4.4.2, "Task identification," listing 10 basic activities. The 2025 version merges these concepts and elevates them into a new section, 4.2, "Characteristics of robotic applications and robot cells." This is no longer a layout guide but a mandatory list of characteristics that must be considered in the risk assessment according to ISO 12100. The list of tasks (now in 4.2 e) has been substantially expanded from 10 to 22 items, covering the entire cell life cycle, including commissioning, application changes, software integration, reinstallation, and disposal.

**Summary and impact**

Chapter 4 of ISO 10218-2:2025 has been significantly reformed, aligning with the changes in previous chapters. It has transitioned from being a guide to risk assessment to becoming a specialized supplement to ISO 12100. By delegating the general process while simultaneously specifying mandatory inputs in an unprecedented manner (a list of 22 tasks), it introduces a completely new methodology for evaluating collaborative contact. This places higher process demands on the integrator, who can no longer rely on a simple checklist for layout design but must instead apply the ISO 12100 process to a much broader and technically more demanding set of input characteristics. The most significant changes within this part of the standard are clearly summarized in the following Table 8.

**Table 8** Key changes in risk assessment methodology in ISO 10218-2 (2011 vs. 2025)

| Area | ISO 10218-2:2011 [4] | ISO 10218-2:2025 [3] | Changes and Implications | Category |
|---|---|---|---|---|
| Methodological Approach | The chapter described the risk assessment process, essentially duplicating the principles of ISO 12100. | The chapter delegates the entire risk assessment process to ISO 12100. The remainder of the chapter consists of informative notes and specific inputs. | **Change:** The standard ceases to be a guide and becomes a specialized supplement to ISO 12100. **Implication:** The integrator must fully apply the process described in ISO 12100. | Structural and methodological changes |
| Task Identification | Section 4.4.2 defines 10 basic tasks focused primarily on operation, setting, and maintenance. | Section 4.2 e) expands the list to 22 items. Newly covers the entire life cycle. | **Change:** Increase in analysed tasks. **Implication:** Risk assessment must newly cover life cycle phases that were not previously required. | Structural and methodological changes |
| Contact Risk Assessment | Methodology for evaluating human-robot contact is entirely absent. | New Section 4.3.2 introduces a specific methodology for contact assessment. | **Change:** Addition of methodology for biomechanical risk assessment. **Implication:** The integrator must newly analyse force, pressure, and contact type. | Technical changes |
| Chapter Structure and Title | Title "Hazard identification and risk assessment". Chapter is divided into 5 main sections describing the process (Why, What, How). | Title "Risk assessment". Chapter is divided into 3 main sections describing delegation, general inputs, and special inputs. | **Change:** Title simplification and restructuring. **Implication:** New structure reflects the philosophy of delegating the general process and specifying unique characteristics. | Structural and methodological changes |

*4.4. Analysis of safety requirements and measures*

Chapter 5 constitutes the core component of the standard, defining specific technical measures for risk mitigation in robotic applications. Between the 2011 and 2025 editions, there has been a substantial expansion of content, restructuring, and deepening of technical detail. While the 2011 version was organized into 12 subchapters, the 2025 version comprises 16, many of which are newly introduced or significantly expanded.

**New chapter organization**

The 2025 edition consistently applies the shift in terminology and philosophy. The chapter title has changed from "Safety requirements and protective measures" (2011) to "Safety requirements and risk reduction measures" (2025), aligning more closely with the terminology of ISO 12100. A significant structural difference is the introduction of new sections focused on design:

- **Design (5.2):** The 2025 edition introduces a completely new, detailed section 5.2, specifying requirements for materials, mechanical strength, lifting, and stability. These requirements were mentioned only marginally or not at all in the 2011 version.

- **Cybersecurity (5.2.16):** This is a completely new requirement mandating a cyber threat assessment and the implementation of measures against unauthorized access to control systems. This topic was absent in the 2011 edition.

**Functional safety and control systems**

The domain of functional safety (previously 5.2, now 5.5) has also undergone significant modernization.

- **2011 Edition:** Generally required PL=d / SIL 2 for safety functions, with the possibility of deviation based on risk assessment.

- **2025 Edition:** Is considerably more specific. It introduces detailed requirements for the parameterization of safety functions (5.5.5), including checksum generation and configuration verification. It specifies requirements for safe communication (5.5.9) and defines network categories (1-3) according to IEC 61508. Specific safety functions such as start/restart interlock (5.5.7), monitored standstill (5.5.8), and monitored speed (5.5.6.2), which were not explicitly codified in the previous version, are newly defined.

**Collaborative applications**

The most significant addition is Section 5.14, "Collaborative applications."

- **2011 Edition:** Contained only a brief section 5.11, listing four collaboration methods (safety-rated monitored stop, hand guiding, speed and separation monitoring, power and force limiting) but provided no details for their implementation.

- **2025 Edition:** Elaborates on these methods in detail. For the Power and Force Limiting (PFL) method (5.14.6), it introduces a methodology for assessing quasi-static and transient contact, defines requirements for reducing the risk of contact events (e.g., edge rounding, increased surface area, soft surfaces), and references the new Annexes M and N for biomechanical limits and measurements.

**Operating modes and control**

The section dedicated to operating modes (5.7) has been significantly revised to reflect modern control methods.

- **High-speed (5.7.2.3.3):** This is newly defined as an optional mode intended exclusively for program verification (not production), requiring the mandatory use of a 3-position enabling device and compliance with other safety conditions.

- **Remote control (5.7.6.3):** The new standard addresses the issue of remote access and control (e.g., via a network), defining activation conditions and the priority of local control.

- **Cableless or detachable pendants (5.7.8.4):** Specific requirements for cableless teach pendants are introduced, including pairing, signal loss detection, and unique identification of the controlled robot.

**Summary and impact**

Chapter 5 of ISO 10218-2:2025 represents an innovative tool for modern robot integration. Compared to the 2011 version, which focused on basic physical safety and stop logic, the new edition places fresh emphasis on software defined safety, collaborative interaction, and cyber resilience. The most significant changes within this part of the standard are summarized in the following Table 9.

**Table 9** Key modifications in safety requirements and protective measures in ISO 10218-2 (2011 vs. 2025)

| Area | ISO 10218-2:2011 [4] | ISO 10218-2:2025 [3] | Changes and Implications | Category |
|---|---|---|---|---|
| Collaborative Applications | Brief mention in 5.11. Lists methods but lacks technical details for power and force limiting. | Extensive Section 5.14. Defines quasi-static/transient contact and measures for contact risk reduction. | **Change:** Definition of contact types. **Implication:** Enables certification of collaborative applications. | Technical changes |
| Cybersecurity | Topic is not mentioned in Chapter 5. | New Section 5.2.16 requiring threat assessment and measures against unauthorized access to control. | **Change:** Introduction of requirements for a new type of security. **Implication:** The integrator must address network security and access rights. | Technical changes |
| Safety Parameterization | Requires only general control system performance (PL/SIL). | New Section 5.5.5 defines requirements for software configuration, checksums, and protection of parameters against unauthorized change. | **Change:** Standardization of safety software management. **Implication:** Increases demands on software validation. | Technical changes |
| Remote and Cableless Control | Contains only basic requirements for control location (5.3.2). | Detailed Sections 5.7.6 and 5.7.8.4 address network control and cableless teach pendants. | **Change:** Adaptation to modern control technologies. **Implication:** Clear rules for control priority. | Technical changes |
| Mechanical Design | General requirements for hazard avoidance. | New Section 5.2.2 specifies requirements for materials, strength (factors 1.25/1.1), and mechanical design of application parts. | **Change:** Refinement of engineering requirements for the entire application. | Technical changes |

| | | | **Implication:** Increases emphasis on structural integrity of tools, fixtures, and frames in the cell. | |
| Motion Limiting | Mentions "soft axis" as an option in 5.4.3. | Section 5.4.7 details mechanical, non-mechanical, and dynamic limiting. | **Change:** Deeper integration of software limits. **Implication:** Enables more flexible cell design while maintaining safety. | Technical changes |
| End-effectors | Section 5.3.10 focuses mainly on energy loss and load retention. | Expanded Section 5.9 addresses surface shapes, tool change systems, and specific gripping risks. | **Change:** More detailed design requirements. **Implication:** End-effector risk assessment is now bound to new, specific standard regulations. | Technical changes |
| Speed Monitoring | Mentioned in the context of manual mode (<250 mm/s). | Defined as a specific safety function (5.5.6) distinguishing Reduced speed (Max. 250 mm/s) and Monitored speed. | **Change:** Terminological and functional refinement. **Implication:** Enables more sophisticated speed control in different process phases. | Technical changes |

### 4.5. Verification, validation, and information for use

Chapters 6 and 7, which conclude the normative text, have undergone profound restructuring between the 2011 and 2025 editions. What was originally a relatively loose set of recommendations and general references to ISO 12100 has evolved into highly specific, technical instructions that precisely define the integrator's obligations regarding conformity assessment and information transfer to the user.

**Verification and validation**

Chapter 6 in the 2025 edition has been condensed in terms of page count and text volume, accompanied by a notable shift in the binding nature of the accompanying annexes and the introduction of new technical requirements for collaborative applications.

- **2011 Version:** The chapter was structured to define verification methods directly within the text (e.g., visual inspection, practical tests) and referenced Annex G for specific requirements. This annex held the status of a normative (binding) part of the document, implying that the specified list of properties had to be verified strictly in accordance with the standard.

- **2025 Version:** The text of the chapter has been shortened and delegates the methodological procedure to the new Annex H. There is a formal change here, although Annex H contains a detailed validation table that assigns the requirements of Chapter 5 to specific methods—like the previous edition—its status is now merely informative. The verification obligation itself remains strictly anchored in the main text; however, the table in the annex serves as a methodological guide rather than a rigid prescription, affording experts greater flexibility in selecting an adequate validation method.

Furthermore, a specific section 6.3.3, "Biomechanical limits," has been newly introduced, mandating the measurement of forces and pressures during contact events. This constitutes a completely new technical requirement absent in the 2011 edition, necessitating specialized measuring equipment and procedures as described in Annex N.

**Information for use**

Chapter 7 has been expanded in terms of both scope and detail. From the original three subchapters in 2011, the chapter now comprises five sections, with Subchapter 7.5, "Instruction handbook," expanding from 10 to 23 detailed subsections.

- **Harmonization with ISO 20607:** The 2025 edition (in 7.5.1) now normatively requires that the instruction handbook complies with the structure defined in ISO 12100:2010 and ISO 20607:2019. This unifies the documentation format across the industry.

- **Detailed technical data:** While the 2011 version required rather general descriptions, the 2025 version demands the provision of precise technical data. For instance, Section 7.5.16, "Functional safety," requires a detailed specification of all safety functions, including their PL/SIL, inputs/outputs, and response times. Section 7.5.22, "Collaborative applications," mandates the specification of biomechanical limits, contact areas, and effective masses for PFL applications.

- **Cybersecurity:** A completely new Section 7.5.23 requires the integrator to provide the user with information regarding implemented cybersecurity functions (firewalls, recovery plans, etc.).

- **Digital marking:** Section 7.3 now (consistent with Part 1) permits the use of digital identifiers (e.g., QR codes) to make extensive information available that cannot be accommodated on a physical label.

**Summary and impact**

The transformation of Chapters 6 and 7 signifies a substantial increase in the administrative and technical burden for integrators. Validation now requires rigorous mapping in accordance with the table in Annex H. In the domain of collaborative applications, the standard establishes an obligation to measure biomechanical quantities. Consequently, the documentation (Instruction handbook) evolves from a simple user manual into a comprehensive technical document that must contain detailed data on functional safety, biomechanics, and cyber resilience, structured according to ISO 20607. The most significant changes within this part of the standard are clearly summarized in the following Table 10.

**Table 10** Key changes in validation methodologies and documentation requirements in ISO 10218-2 (2011 vs. 2025)

| Area | ISO 10218-2:2011 [4] | ISO 10218-2:2025 [3] | Changes and Implications | Category |
|---|---|---|---|---|
| Structure of Chapter 7 | Contains 10 subsections. Requirements are rather general. | Contains 23 subsections. Newly covers topics such as cybersecurity, PFL applications, SSM, response times. | **Change:** The chapter is more detailed and extensive. **Implication:** The instruction handbook becomes a comprehensive technical document. | Structural and methodological changes |
| Normative Framework for Documentation | References ISO 12100 and IEC 60204-1 generally. | Normatively requires compliance with ISO 20607:2019 (Instruction handbook). | **Change:** Standardization of handbook structure. **Implication:** Unifies the form of documentation. | Structural and methodological changes |

| Functional Safety Documentation | Requires specification of safety requirements and integrity (7.2.5 c). | New Section 7.5.16 requires detailed data: intended function, PL/SIL, inputs/outputs, response times, and parameter settings. | **Change:** Requirement to provide complete engineering safety data. **Implication:** The user receives data necessary for safety life cycle management, not just a function description. | Technical changes |
|---|---|---|---|---|
| Collaborative Applications | Mentions only a statement of suitability for collaboration (7.2.10). | New Section 7.5.22 requires specific data for PFL (masses, contact areas, biomechanical limits) and SSM (number of persons, stopping distances). | **Change:** Introduction of specific information obligations for collaborative systems. **Implication:** The user must be informed in detail about collaboration limits and residual risks. | Technical changes |
| Cybersecurity | Topic is entirely absent from the standard. | New Section 7.5.23 requires information on implemented functions (firewalls) and necessary user measures (updates). | **Change:** Introduction of information obligations regarding IT security. **Implication:** The integrator must transparently communicate implemented measures and expectations for the user to maintain security. | Technical changes |
| Marking | Requires physical marking (7.3). | Permits the use of digital identifiers (QR code) to provide complete data if space is limited. | **Change:** Modernization of marking. **Implication:** Enables access to digital documentation directly from the machine. | Technical changes |

## 4.6. Analysis and restructuring of annexes

The system of annexes in ISO 10218-2 has undergone a significant expansion in the 2025 edition. From the original eight annexes, the collection has grown into a comprehensive set extending up to the letter Q. Although most of these annexes remain formally informative, their content has shifted qualitatively. They no longer serve merely as general guidelines but provide detailed technical materials offering specific methodologies for calculations, measurements, and validation of applications, thereby significantly facilitating the practical implementation of requirements from the main text of the standard.

**Updating and relocation of existing annexes**

Several original annexes have been retained but significantly modernized:

- **List of significant hazards (Annex A):** While remaining informative, it has been expanded to include a second table specifically focused on hazards associated with end-effectors.

- **Safeguarding material entry and exit points (Annex C vs. K):** The original informative Annex C, addressing conveyors and tunnels, has been transformed into the new Annex K. The content has been modernized and supplemented with more detailed diagrams and principles for protection against whole-body access.

- **Relationship to other standards (Annex B vs. Q):** The original Annex B has been relocated to the end of the document as Annex Q and updated with references to new standards.
- **Harmonization (Annex ZA):** This annex has been moved to the beginning of the document and transformed into a clear table mapping specific clauses of the standard to the requirements of the Directive.

**Change in verification status (G vs. H)**

A significant formal change has occurred in verification. The original Annex G (2011), which defined the means of verification, held normative status (i.e., mandatory). In the new edition, it is superseded by Annex H (2025). Although this new annex is substantively more detailed and offers a similar table linking requirements to verification methods, its status has been changed to informative. The binding nature of verification is now fully de-fined within the text of the standard, while the annex serves as a methodological guide.

**New technical and methodological annexes**

The most significant contribution of the 2025 edition is the introduction of entirely new annexes that had no equivalent in 2011:

- **Functional safety (Annex C and D):** The new normative Annex C defines the required performance levels (PL/SIL) for safety functions. The informative Annex D subsequently provides a standardized format for their documentation.
- **Collaborative and biomechanical limits (Annex M and N):** A completely new Annex M has been introduced, defining biomechanical limits for quasi-static and transient contact (pressure and force on various body parts). This is followed by Annex N, which describes the methodology for measuring these quantities using specialized measuring devices.
- **Specific applications (Annex I and J):** The new Annex I addresses risks and the design of end-effectors (grippers) in detail. Annex J provides guidance on safeguarding manual loading and unloading stations, including the prevention of bypassing safeguards.
- **Illustration of spaces (Annex B):** A new graphical annex visualizing the definitions of maximum, restricted, and protected spaces, thereby unifying the interpretation of these concepts.

**Summary and impact**

The expansion of annexes in ISO 10218-2:2025 is a direct consequence of the introduction of new requirements in the main text of the document. The annexes now focus to a greater extent on defining specific workflows and methodologies, thus fulfilling the function of a technical manual for verification and calculations, while retaining their original role as recommendations for proper integration. Integrators can now find precise instructions for calculating separation distances, methodologies for measuring contact forces, and tables for determining safety performance levels. Although most annexes, including the validation annex, remain informative, their technical depth and coherence with the main text provide a solid framework for the correct application of the standard. This enhances the capability of integrators to implement and verify even complex applications based on standardized procedures, without the standard necessarily dictating a single possible solution. The most significant changes within this part of the standard are clearly summarized in the following Table 11.

**Table 11** Key changes in the Annexes of ISO 10218-2 (2011 vs. 2025)

| Area | ISO 10218-2:2011 [4] | ISO 10218-2:2025 [3] | Changes and Implications | Category |
|---|---|---|---|---|
| List of Significant Hazards (Annex A – informative) | Informative list of 10 risk groups derived from ISO 12100. | Expanded to two tables. Newly contains a specific hazard table for end-effectors. | **Change:** More detailed hazard specification. **Implication:** Risk analysis of the tool (gripper) itself. | Technical changes |
| Verification and Validation (G – normative vs. H – informative) | Table for verification and validation of requirements in Chapter 5. | Significantly expanded table for verification and validation of requirements in Chapter 5. | **Change:** Change of status from normative to informative. **Implication:** Binding nature is in the main text; the annex provides flexibility in method selection. | Structural and methodological changes |
| Material Entry (C – informative vs. K – informative) | Basic drawings of tunnels and guards with descriptions. | Updated drawings and principles for conveyors and openings. | **Change:** Modernization and redesignation of the annex. **Implication:** Retains proven protection principles for conveyors in an updated form. | Structural and methodological changes |
| Functional Safety PL/SIL (Annex C – normative) | Did not exist. Addressed generally in the text. | Table defining minimum required PL/SIL for specific safety functions. | **Change:** Introduction of minimum binding safety levels. **Implication:** Eliminates ambiguity in determining PL for standard functions. | Technical changes |
| Biomechanical Limits (Annex M) | Did not exist. | Defines pressure and force limits for 29 body regions (quasi-static/transient contact). | **Change:** Introduction of biomechanical limits into the standard. **Implication:** Provides data necessary for PFL application design. | Technical changes |
| Force and Pressure Measurement (Annex N – informative) | Did not exist. | Methodology for measuring contact events and calibrating measuring devices. | **Change:** Standardization of measurement methodology. **Implication:** Defines exactly how to verify compliance with biomechanical limits. | Technical changes |
| Manual Loading (Annex J – informative) | Did not exist. | Detailed guide for safeguarding manual stations (prevention of bypassing, intrusion detection). | **Change:** Specific guide for a frequent application type. **Implication:** Addresses the issue of defeating safeguards and ergonomics at operator interaction points. | Technical changes |
| Illustration of Spaces (Annex B – informative) | Did not exist. | Graphic schemes visualizing maximum, restricted, operating, and protected spaces. | **Change:** Visualization of definitions. **Implication:** Improves understanding of relationships between different zones and protective device placement. | Structural and methodological changes |

### 4.7. Integration of ISO/TS 15066:2016 into ISO 10218-2:2025

The technical specification ISO/TS 15066:2016 [6], which served for nearly a decade as the de facto standard for the design and validation of applications involving shared human-robot workspaces, has been fully absorbed into the structure of ISO 10218-2:2025 [3].

Through this integration, methodologies previously regarded as supplementary or recommended have become mandatory components of the fundamental safety regulation for robot integration.

**Transformation of the Power and Force Limiting (PFL) methodology**

The most significant aspect of this integration involves the transfer of the methodology for Power and Force Limiting (PFL) applications. In 2016, ISO/TS 15066 first defined biomechanical limits of the human body for robotics purposes. In the new edition of ISO 10218-2:2025, this concept is elevated to a normative requirement in Section 5.14.6. The standard adopts the distinction between contact events originally introduced by TS 15066:

- **Quasi-static contact:** A clamping or crushing situation where a body part is constrained and cannot recoil.
- **Transient contact:** A dynamic impact where the affected body part can recoil or retract following the impact.

While TS 15066 defined these terms in Chapter 3, the new standard integrates them directly into the terminology section (3.1.12) and subsequently utilizes them within the risk assessment methodology (4.3.2). The original Annex A of TS 15066, which contained tables of pain thresholds for 29 body regions, has been transformed into Annex M (in-formative) in ISO 10218-2:2025. Although it retains its informative status, the reference in the main text (5.14.6.1) renders it an essential reference for configuring safety function parameters.

**Speed and Separation Monitoring (SSM)**

The Speed and Separation Monitoring (SSM) methodology, described in Chapter 5.5.4 of TS 15066, has been significantly elaborated in the new standard and anchored in Section 5.14.5. The original formulas for calculating the protective separation distance, which accounted for system reaction times and operator speeds, have been relocated to the new normative Annex L.

A significant advancement is that the new standard more explicitly links SSM with functional safety requirements—an aspect not addressed to such depth in the original technical specification.

**Validation and measurement**

The technical specification ISO/TS 15066 pioneered the definition of a methodology for measuring forces and pressures using specific pressure and force measurement devices (PFMD). This section (originally Annex B in ISO/TS 15066), representing an approach critically analysed in recent biomechanical studies [14,15], has been revised into Annex N (informative) within the new standard. The integration of this annex provides integrators with clear guidance on the practical verification of compliance with biomechanical limits, defining parameters for damping materials, spring stiffness of measuring instruments, and measurement procedures for worst-case scenarios.

**Summary of integration**

The incorporation of ISO/TS 15066 into ISO 10218-2:2025 resolves the situation where integrators were compelled to combine the requirements of a general standard with a separate technical specification to ensure compliance. Collaborative applications are no longer viewed as exceptions but as standard components of the robotic application spectrum. Concepts such as PFL and SSM are now integral parts of the standard's vocabulary, and their implementation is subject to the same rigorous verification and validation requirements as any other safety function. The integration of the technical specification into the second part of ISO 10218-2:2025 is summarized in the following Table 12.

**Table 12** Transposition of ISO/TS 15066 requirements into the structure of ISO 10218-2:2025

| Area | ISO 10218-2:2011 [3] | ISO/TS 15066:2016 [6] | Changes and Implications |
|---|---|---|---|
| Contact Definition | Chapter 3.4 (Quasi-static), 3.5 (Transient). | Chapter 3.1.12 (Contact – related). | **Full integration.** Definitions have become part of the standard vocabulary, unifying the understanding of collision states. |
| Collaborative Operations (4 methods) | Chapter 5.5 (Monitored stop, Hand guiding, SSM, PFL). | Chapter 5.14 (Collaborative applications) and subsections 5.14.4–5.14.6. | **Expansion and normative anchoring.** Methods are described in greater detail and linked to PL/SIL requirements in Annex C. |
| Contact Risk Assessment | Chapter 4.3 (Hazard identification and risk assessment). | Chapter 4.3 (Characteristics of collaborative applications). | **Systematization.** Newly mandates specific risk assessment for face/head contact (5.14.6.3), previously less emphasized. |
| Biomechanical Limits (Body Model) | Annex A. Limits for quasi-static and transient contact. | Annex M (Informative). Limits for quasi-static and transient contact. | **Content retention.** Values have been retained but are now part of the main standard, increasing their weight in conformity assessment. |
| Measurement Methodology (PFL) | Hinted at in the text; absence of a detailed normative annex for measurement (addressed in the informative part). | Annex N (Informative). Detailed methodology for PFMD, calibration, and measurement procedure. | **Validation standardization.** The standard provides a precise measurement process, including specifications for the measurement chain and sample size. |
| SSM Calculation (Distance) | Chapter 5.5.4. Formulas for calculating minimum distance. | Annex L (Normative). Speed and separation monitoring – separation distance. | **Elevation to normative status.** Formulas for calculating the protective separation distance are now a binding part of the standard (normative annex). |
| Information for Use | Chapter 7. Specific requirements for PFL and SSM documentation. | Chapter 7. Specific requirements for PFL and SSM documentation. | **More detailed requirements.** The new standard requires stating specific technical parameters (effective masses, contact areas) directly in the instruction handbook. |

## 5. Discussion

The analysis of the changes between the 2011 and 2025 editions of ISO 10218-1 and ISO 10218-2, supplemented by a detailed examination of the transposition of the technical specification ISO/TS 15066, reveals several cross-cutting trends. These changes will have a tangible impact on industrial practice, particularly in the areas of design, validation, and operation of robotic workplaces. The following chapter discusses the economic, qualification, and methodological implications of this new normative framework.

**Economic and time impacts on integration: Challenges and opportunities for research**

The introduction of mandatory measurements for collaborative applications (now anchored in the normative framework of ISO 10218-2 [3], Annex N) and the requirement for more detailed validation (Annex G in Part 1, Annex H in Part 2) are likely to extend the time required for commissioning equipment. As highlighted by Valori et al., validating safety in collaborative applications requires rigorous adherence to biomechanical limits, which has traditionally been a complex challenge for integrators [15]. Integrators will need to factor in the costs of specialized measurement equipment (for PFL applications) and the increased administrative burden associated with creating technical documentation in accordance with ISO 20607 and functional safety specifications. It can be assumed that engineering costs will rise even for simpler applications, which may paradoxically slow down the deployment of collaborative robots in cases where collaboration is not strictly necessary.

However, this pressure for accuracy and validation also opens significant scope for applied research in the field of simulation. Since physical tests of collaborative applications require the integrator to expertly select critical measurement points based on experience, there is potential for the utilization of digital twins. Research into the deployment of advanced simulations, which could predict contact forces based on an accurate mathematical model of the robot and its dynamic parameters, could significantly streamline this process.

Simulation software could initially serve to identify critical points where limit forces and pressures are likely to be exceeded, thereby reducing the number of necessary physical measurements. In the long term, the development of reliable simulation models could theoretically lead to a scenario where extensive physical measurements would no longer be required, as they would be replaced by validated software estimates. This is also related to the need to standardize data formats and measurement documentation, an issue already partially addressed in the ISO/PAS 5672:2023 [19].

**Shift in qualification requirements: Cybersecurity as a necessity**

New requirements for cybersecurity and software parameterization (software limits, checksums) are altering the profile of the required expert. Cybersecurity in the context of industrial robotics has not previously been explicitly addressed in standards; however, with the advent of Industry 4.0, it has become a necessity. Recent research emphasizes that connected robots are vulnerable to cyber-attacks that can compromise not only data but also physical safety [10], validating the inclusion of mandatory cybersecurity clauses in the 2025 edition.

Cybersecurity in the context of industrial robotics has not previously been explicitly addressed in standards; however, with the advent of Industry 4.0, defined by high levels of connectivity and data exchange, it has become a necessity. If a robotic network is connected to the Internet or corporate systems, the risk of external attack increases. Although companies have invested in this area previously, it is now a normative requirement. The rationale is not only the protection of sensitive production data but, crucially, the prevention of physical risks. Unauthorized access allowing remote reprogramming, unexpected production stoppages, or the deactivation of safety functions could lead to unpredictable robot movements and a direct threat to human health. The standard thus responds to the fact that, in the modern concept, cybersecurity is an inseparable component of functional safety.

**Legal and liability aspects: Delegation and flexibility**

The decision by standards bodies to fully delegate the risk assessment methodology to ISO 12100 and the instruction handbook structure to ISO 20607 simplifies the text of ISO 10218 but increases the demands on the user's orientation within the system of standards. The integrator must actively work with the entire ecosystem of Type B standards rather than relying on a single robotic standard. From a legal perspective, there is a significant shift in the definition of roles (manufacturer vs. integrator vs. user), which more clearly delineates the boundaries of responsibility.

Conversely, this approach affords greater flexibility in the choice of verification methods. Changing the status of the annex containing the validation table from normative to informative (Annex H in Part 2) means that the integrator is not rigidly bound to a single procedure, although the table remains an important guide. Unifying the risk assessment process with the general standard for machinery (ISO 12100) eliminates duplication and allows integrators to employ a single methodology for the entire robotic workplace, including peripheral machinery, thereby simplifying administrative processes in complex projects.

**Standardization of collaborative applications**

The integration of biomechanical limits sends a clear signal. For applications requiring higher speeds, it will often remain more advantageous to use Speed and Separation Monitoring (SSM) technology, as reviewed in [18], or traditional spatial separation rather than Power and Force Limiting (PFL). At the same time, this integration implies that collaborative applications, and their safety trends discussed in [17], are no longer perceived as a special discipline.

At the same time, however, this integration implies that collaborative applications are no longer perceived as a special or experimental discipline. The standard thus confirms that collaborative robots are a standard component of modern robotics. Integrating the technical specification directly into the ISO 10218-2 standard is a logical step, as it eliminates the need to search for requirements in separate documents and unifies the rules for integrating both classic and collaborative robots within a single standard. This simplifies orientation for integrators and should reinforce the principle that the deployment of a collaborative robot should not be determined solely by current trends, but by the technical feasibility of the application in accordance with clearly defined safety limits.

This section may be divided by subheadings. It should provide a concise and precise description of the experimental results, their interpretation, as well as the experimental conclusions that can be drawn.

## 6. Conclusions

The primary objective of this article was to conduct a comparative analysis of the safety standards ISO 10218-1 and ISO 10218-2 in their 2011 and 2025 editions, supplemented by an evaluation of the integration of the technical specification ISO/TS 15066. The findings confirm that the 2025 revision represents not merely an evolutionary update but a generational overhaul of the normative framework, reflecting a decade of technological advancement in automation.

While the 2011 edition was primarily focused on the definition of safe hardware, the 2025 edition shifts the focus toward the safety of the application and process. This shift is most evident in the field of collaborative robotics. The direct absorption of methodologies from ISO/TS 15066:2016 into the text of ISO 10218-2:2025 concludes a phase where collaborative applications were perceived as a niche discipline peripheral to the mainstream. The standard now affirms that human-robot collaboration is an integral component of modern manufacturing, albeit one subject to strict physical and validation limits.

In many respects, the new standard merely codifies the status quo demanded by the reality of Industry 4.0, while simultaneously paving the way for Industry 5.0. The emphasis on safe and close human-robot cooperation is a necessary technical prerequisite for the human-centric approach to production advocated by this upcoming phase of industrial evolution [20].

From a commercial perspective, the global nature of the new ISO standard offers a significant advantage. The unification of technical requirements at the international level eliminates regional barriers and facilitates the export of technologies to markets outside the European Union for manufacturers and integrators, without the need to navigate fragmented local regulations.

Although the scope of changes—ranging from mandatory measurements and cyber-security to documentation—may appear drastic at first glance, the actual market impact is unlikely to be as disruptive. Leading manufacturers and advanced integrators have monitored the standard's development since the preparatory stages and have already de facto implemented many of the now mandatory requirements into their products and processes based on available drafts and technical trends. Consequently, the standard often validates and standardizes existing best practices rather than creating barriers that prepared companies would be unable to overcome.

The ISO 10218 standard in the 2025 edition thus ceases to be a mere list of design constraints and transforms into a comprehensive engineering manual. While this manual demands high levels of expertise, in return, it offers more robust tools for risk management in dynamic environments, opens the door to advanced verification methods using simulation, and ensures that safety standards serve not as a hindrance, but as a guarantor of stability for the continued advancement of robotics.

**Author Contributions:** Conceptualization, D.H. and K.H.; methodology, D.H.; validation, D.H., A.V., K.H., V.L. and A.B.; Investigation, D.H and K.H.; resources, D.H. and K.H.; writing—original draft preparation, D.H.; writing—review and editing, A.V., K.H. and V.L.; supervision, A.V.; project administration, A.V.; funding acquisition, A.B. All authors have read and agreed to the published version of the manuscript.

**Funding:** This research was funded by Czech Republic and the European Union within the framework of the project REFRESH – Research Excellence For REgion Sustainability and High-tech Industries (No. CZ.10.03.01/00/22_003/0000048) through the Fair Trans-Formation Operational Program.

This work was supported by the European Regional Development Fund under the project Research Platform for Digital Transformation and Society 5.0 CZ.02.01.01/00/23_021/0012599 within the Jan Amos Komensky Operational Program.

**Institutional Review Board Statement:** Not applicable. Informed Consent Statement: Not applicable.

**Data Availability Statement:** No new data were created or analyzed in this study. Data sharing is not applicable to this article.

**Acknowledgments:** During the preparation of this work, the authors used Gemini (version 3 Pro) to improve language and readability. After using this tool/service, the authors reviewed and edited the content as needed and take full responsibility for the content of the publication.

**Conflicts of Interest:** The authors declare no conflict of interest.

## References

1    ISO 10218-1:2025. *Robotics – Safety requirements – Part 1: Industrial robots*; International Organization for Standardization: Geneva, Switzerland, 2025.

2    ISO 10218-1:2011. *Robots and robotic devices – Safety requirements for industrial robots – Part 1: Robots*; International Organization for Standardization: Geneva, Switzerland, 2011.

3    ISO 10218-2:2025. *Robotics – Safety requirements – Part 2: Industrial robot applications and robot cells*; International Organization for Standardization: Geneva, Switzerland, 2025.

4    ISO 10218-2:2011. Robots and robotic devices – Safety requirements for industrial robots – Part 2: Robot systems and integration; International Organization for Standardization: Geneva, Switzerland, 2011.

5    ISO 12100:2010. *Safety of machinery – General principles for design – Risk assessment and risk reduction*; International Organization for Standardization: Geneva, Switzerland, 2010.

6    ISO/TS 15066:2016. *Robots and robotic devices – Collaborative robots*; International Organization for Standardization: Geneva, Switzerland, 2016.

7    Krippendorff, K. *Content Analysis: An Introduction to Its Methodology*, 4th ed.; SAGE Publications: Los Angeles, CA, USA, **2019.**

8    Hsieh, H.-F.; Shannon, S.E. Three Approaches to Qualitative Content Analysis. *Qual. Health Res.* **2005**, *15*, 1277–1288. [CrossRef]

9    Tanimu, J.A.; Abada, W. Addressing cybersecurity challenges in robotics: A comprehensive overview. *Cyber Security and Applications* **2025**, *3*, 100074. [CrossRef]

10   Dutta, V.; Zielińska, T. Cybersecurity of Robotic Systems: Leading Challenges and Robotic System Design Methodology. *Electronics* **2021**, *10*, 2850. [CrossRef]

11   ISO 20607:2019. *Safety of machinery – Instruction handbook – General drafting principles*; International Organization for Standardization: Geneva, Switzerland, 2019.

12   ISO 13849-1:2023. Safety of machinery – Safety-related parts of control systems – Part 1: General principles for design; International Organization for Standardization: Geneva, Switzerland, 2023.

13   IEC 62061:2021. *Safety of machinery - Functional safety of safety-related control systems*; International Electrotechnical Commission: Geneva, Switzerland, 2021.

14   Samarathunga, S.M.B.P.B.; Valori, M.; Legnani, G.; Fassi, I. Assessing Safety in Physical Human–Robot Interaction in Industrial Settings: A Systematic Review of Contact Modelling and Impact Measuring Methods. *Robotics* **2025**, *14*, 27. [CrossRef]

15   Valori, M.; Scibilia, A.; Fassi, I.; Saenz, J.; Behrens, R.; et al. Validating Safety in Human–Robot Collaboration: Standards and New Perspectives. *Robotics* **2021**, *10*, 65. [CrossRef]

16   Zimmermann, J.; Huelke, M.; Clermont, M. Experimental Comparison of Biofidel Measuring Devices Used for the Validation of Collaborative Robotics Applications. *Int. J. Environ. Res. Public Health* **2022**, *19*, 13657. [CrossRef]

17   Arents, J.; Abolins, V.; Judvaitis, J.; Vismanis, O.; Oraby, A.; et al**.** Human–Robot Collaboration Trends and Safety Aspects: A Systematic Review. *J. Sens. Actuator Netw.* **2021**, *10*, 48. [CrossRef]

18   Scholz, C.; Cao, H.-L.; Imrith, E.; Roshandel, N.; Firouzipouyaei, H.; et al. Sensor-Enabled Safety Systems for Human–Robot Collaboration: A Review. *IEEE Sens. J.* **2025**, *25*, 65–88. [CrossRef]

19   ISO/PAS 5672:2023. Robotics – Collaborative applications – Test methods for measuring forces and pressures in human-robot contacts; International Organization for Standardization: Geneva, Switzerland, 2023.

20   Yitmen, I.; Almusaed, A. Synopsis of Industry 5.0 Paradigm for Human-Robot Collaboration. In *Industry 4.0 Transformation Towards Industry 5.0 Paradigm - Challenges, Opportunities and Practices*; IntechOpen: London, UK, **2024**. [CrossRef]